\documentclass[nonacm,natbib=false,sigconf]{acmart}
\usepackage{graphicx}
\usepackage{caption,subcaption}
\usepackage[para]{footmisc}
\usepackage{booktabs}
\usepackage{multirow}
\usepackage{xcolor,colortbl}
\usepackage{hyperref}
\usepackage{cleveref}
\usepackage{todonotes}

\usepackage[style=ACM-Reference-Format,backend=bibtex,sorting=none]{biblatex}
\begin{document}
\title[Multimodal Hate Speech Detection in Bengali]{Multimodal Hate Speech Detection from Bengali\\ Memes and Texts}

\author{Md. Rezaul Karim}
\affiliation{%
  \institution{Fraunhofer FIT \& RWTH Aachen University, Germany}
  \country{}
  }

\author{Sumon Kanti Dey}
\affiliation{%
  \institution{Noakhali Science and Technology University, Bangladesh}
  \country{}}
  
\author{Tanhim Islam}
\affiliation{
  \institution{RWTH Aachen University, Germany}
  \country{}}
  
 \author{Md. Shajalal}
\affiliation{
  \institution{Fraunhofer FIT \& University of Siegen, Germany}
  \country{}}
   
\author{Bharathi Raja Chakravarthi}
\affiliation{%
  \institution{University of Galway, Ireland}
  \country{}}  
  
\renewcommand{\shortauthors}{Karim, Sumon, and Tanhim et al., ``Multimodal Hate Speech Detection in Bengali''}

\begin{abstract}
  Numerous machine learning~(ML) and deep learning~(DL)-based approaches have been proposed to utilize textual data from social media for anti-social behavior analysis like cyberbullying, fake news detection, and identification of hate speech mainly for highly-resourced languages such as English. However, despite of having a lot of diversity and millions of native speakers, some languages like Bengali are under-resourced, which is due to lack of computational resources for natural language processing~(NLP). Similar to other languages, Bengali social media contents also include images along with texts~(e.g., multimodal memes are posted by embedding short texts into images on Facebook). Therefore, only the textual data is not enough to judge them since images might give extra context to make a proper judgement. This paper\footnote{\scriptsize{This is the pre-print version of our accepted and presented paper at International Conference on Speech \& Language Technology for Low-resource Languages~(SPELLL'2022).}} is about hate speech detection from multimodal Bengali memes and texts. We prepared the only multimodal hate speech dataset for-a-kind of problem for Bengali, which we use to train state-of-the-art neural architectures~(e.g., Bi-LSTM/Conv-LSTM with word embeddings, ConvNets + pre-trained language models, e.g., monolingual Bangla BERT, multilingual BERT-cased/uncased, and XLM-RoBERTa) to jointly analyze textual and visual information for hate speech detection\footnote{\textcolor{blue}{\url{https://github.com/rezacsedu/Multimodal-Hate-Speech-Bengali}}}. Conv-LSTM and XLM-RoBERTa models performed best for texts, yielding F1 scores of 0.78 and 0.82, respectively. As of memes, ResNet-152 and DenseNet-161 models yield F1 scores of 0.78 and 0.79, respectively. As of multimodal fusion, XLM-RoBERTa + DenseNet-161 performed the best, yielding an F1 score of 0.83. Our study suggests that text modality is most useful for hate speech detection, while memes are moderately useful. 
\end{abstract}

\maketitle

\section{Introduction}
The micro-blogging sites and social media not only empower freedom of expression and individual voices, but also tempts people to express anti-social behavior~\cite{karim2021deephateexplainer,karim2020classification}, like cyberbullying, online rumours, and spreading hatred statements~\cite{ribeiro2018characterizing}.
Abusive speech expressing prejudice towards a certain group is also very common, and based on race, religion, and sexual orientation is getting pervasive.
United Nations Strategy and Plan of Action on Hate Speech~\cite{guterres2019united} defines hate speech as \emph{any kind of communication in speech, writing or behaviour, that attacks or uses pejorative or discriminatory language regarding a person or a group based on their religion, ethnicity, colour, gender or other identity factors}.

Bengali is one of the major languages in the world. It is spoken by 230 million people in Bangladesh and India. Similar to other major languages such as English, anti-social behaviors like propagating hate speech is also getting rampant in Bengali. Hate speech in Bengali not only signify how severe Bengali hateful statements could be, but also show that hate speech is contextualized from the personal to religious, political, and geopolitical levels~\cite{karim2020classification}. Since such hate speech is getting more pervasive, there is a potential chance that these could lead to serious consequences such as hate crimes~\cite{karim2020classification}, regardless of languages, geographic locations, or ethnicity. 
Addressing and identifying hate speech does not mean limiting or prohibiting freedom of speech, rather keeping it from escalating to a more dangerous level~\cite{guterres2019united}. 

\begin{figure*}
    \centering
    \includegraphics[width=0.7\textwidth]{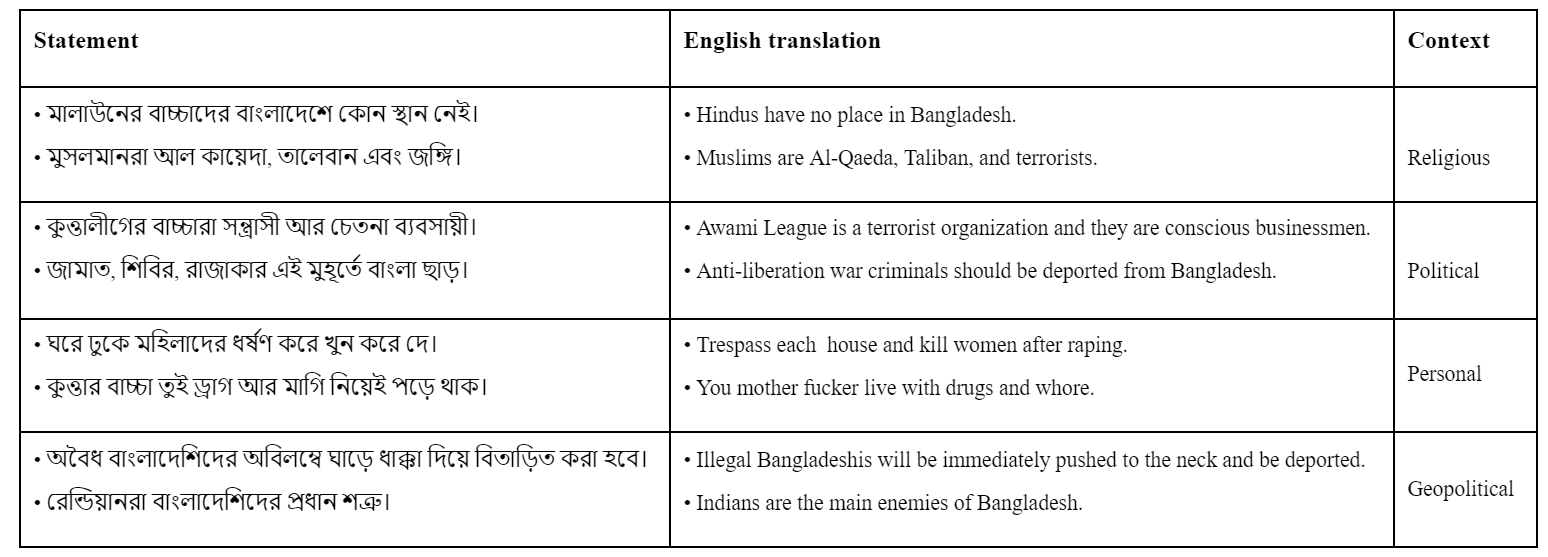}
    \caption{Examples hate speech in Bengali that are either directed towards a specific person, entity, or towards a group~\cite{karim2020classification}} 
    \label{cdec_wf3}
\end{figure*}

\begin{figure*}[h]
    \centering
    \includegraphics[width=0.5\textwidth]{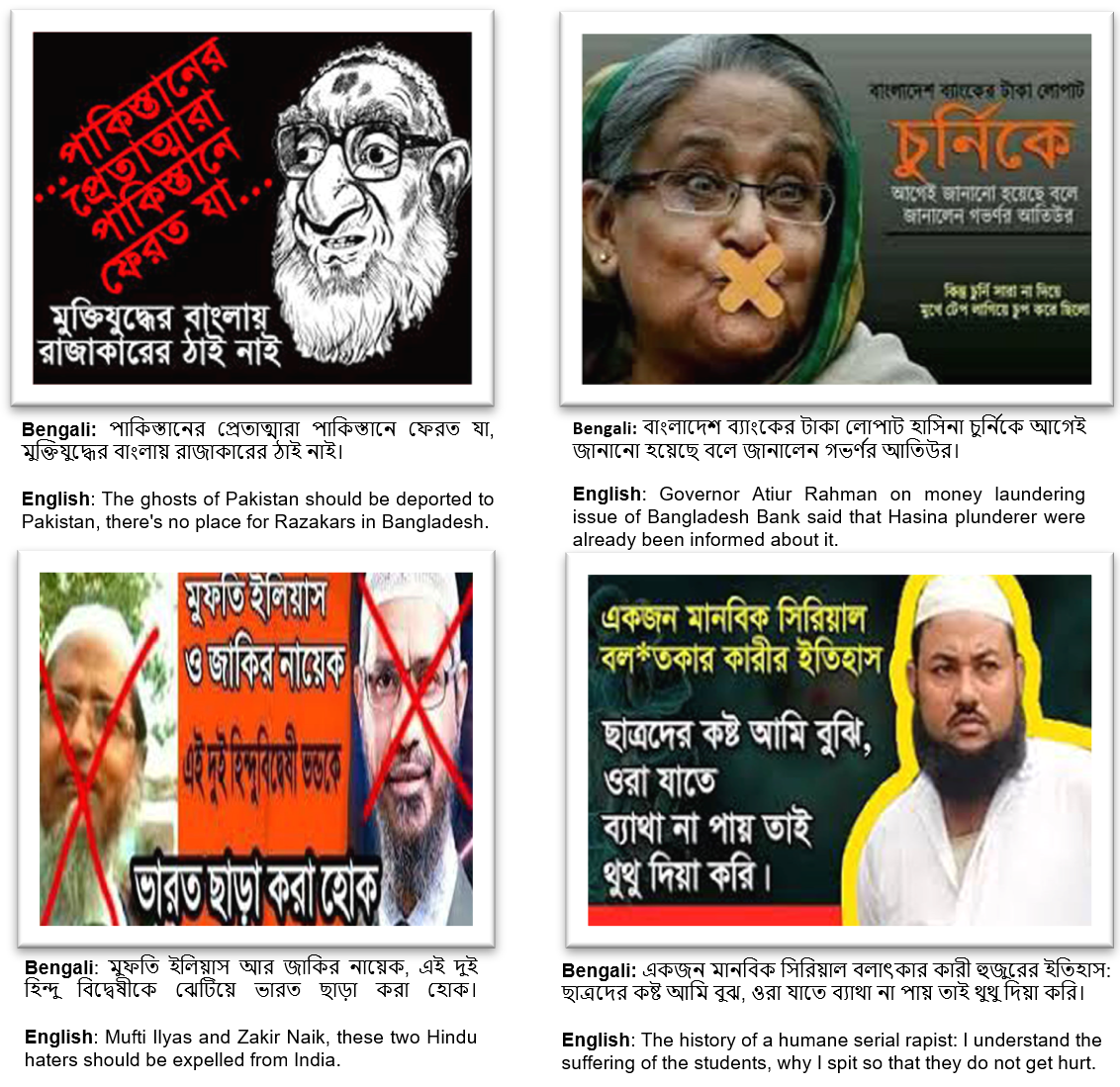}
    \caption{Bengali multimodal memes, where texts and visual information add relevant context for the hate speech detection}    
    \label{fig:multi_meme_example}
\end{figure*}

Automatic identification of such hate speeches in social media and raising public awareness is of utmost importance for Bengali too~\cite{karim2020classification}. Owing to unrestricted access, digitalization, and use of social media, we have access to a huge amount of online contents. Manual reviewing and verification of such large scale data is not only time-consuming and labor-intensive, but also prone to potential human errors~\cite{hate5}. Therefore, robust and scalable methods are required for accurate identification of hate speech in an automated way. Further, similar to other major language, recent Bengali social media content also include images along with texts~(e.g., in Twitter, multimodal tweets are formed by images with short texts embedded into). As shown in \cref{fig:multi_meme_example}\footnote{\scriptsize{\textbf{Disclaimer}: memes and lexicons contain contents that are racist, sexist, homophobic, and offensive in different ways. Further, authors want to clarify that the dataset is collected and annotated from social media for research purposes only and not intended to hurt or offense any specific person, entity, or religious/political groups/parties.}}, some of these multimodal contents are only hate speech because of the combination of the text with a certain image. The presence of only offensive terms does not signify the traces of hate speech, as the hate propagators tend to intentionally construct contents, where the text is not enough to determine they are really hate speech~\cite{hate5}. In such cases, images may provide extra context to make a proper judgement.

Numerous works that have been proposed to accurately identify hate speech in major languages like English that are based on machine learning~(ML) and neural networks~(DNNs)~\cite{elsherief2018hate,hate5}. However, Bengali is severely low-resourced for natural language processing~(NLP), albeit it is a rich language with a lot of diversity. One of the primary reasons is it lacks computational resources such as language models, properly annotated and labelled datasets, and efficient ML methods. Further, similar to other major languages, modern Bengali data consist of texts, images, and memes containing texts that could provide extra context and eventually improve the performance of identifying hate. State-of-the-art~(SotA) transformer language models are becoming increasingly effective in various NLP tasks. Besides, multimodal ML is also being increasingly applied to handle multimodal contents. However, none of the existing works focus on jointly analyzing textual and visual information for hate speech detection in Bengali.   

Inspired by the success of transformer-based language models and SotA multimodal ML learning, we propose
a novel approach for accurate identification of hate speech from Bengali memes and texts. 
In our approach, Bengali memes and texts are first comprehensively preprocessed, before classifying them into either hate speech or neutral w.r.t political, personal, geopolitical, and religious contexts. We apply different multimodal learning techniques in combination with early- and late fusion techniques, including transformer-based neural architectures~(i.e., Bangla BERT-base, multilingual BERT~(mBERT), and XLM-RoBERTa) and CNN architectures~(i.e., VGG, ResNet, DenseNet, and EfficientNet) for the text and imaging modality, respectively. We carried out a wide range of experiments by training several ML and DNN baseline models on which we provide comparative analysis. 

Overall contributions of our paper is 4-folds:

\begin{enumerate}
    \item We prepared the largest and only multimodal hate speech detection dataset to date for the Bengali language.
    \item We train state-of-the-art neural architectures to accurate identification of hateful statements from memes and texts.
    \item Comprehensive evaluation and comparison with baselines.
    \item To foster reproducible research, we make available computational resources such as annotated dataset, language models, and source codes that will further advance the NLP research for under-resourced Bengali language.
\end{enumerate}

The rest of the paper is structured as follows: \Cref{sec:rw} reviews related works. \Cref{sec:methods} describes our approach. \Cref{sec:exp} reports experiment results, showing a comparative analysis with baselines. \Cref{sec:con} summarizes this research, identifies potential limitations and points to some outlooks before concluding the paper.

\section{Related Work}\label{sec:rw}
Numerous works have been proposed to accurately identify hate speech in major languages like English~\cite{elsherief2018hate,hate5}. Many classic methods traditionally rely on manual feature engineering, e.g., support vector machines~(SVM), Na{\"i}ve Bayes~(NB), logistic regression~(LR), decision trees~(DT), random forest~(RF), gradient boosted trees~(GBT). Waseem et al.~\cite{waseem2016hateful} use classical ML techniques to classify 16,000 tweets as racist and sexist. They used word n-grams and character n-grams in conjunction with other task-specific engineered features like gender information and location information. Davidson et al.~\cite{davidson2017automated} used LR along with L1 regularization to decrease the dimensionality of data. They have experimented with multiple traditional ML classifiers like RF, LR, NB, DT, and linear SVMs, where LR and linear SVMs perform significantly better than other models. Malmasi et al.~\cite{zampieri2019semeval} obtained 78\% accuracy in identifying posts across three classes - hate, offensive, and neutral using an approach based on character n-grams, word n-grams, and word skip-grams along with SVM for multiclass classification. 

DNN-based approaches that can learn multi-layers of abstract features from raw texts that are primarily based on convolutional neural networks~(CNNs) or long short-term memory~(LSTM) networks. In comparison with DNN-based methods, ML-based approaches are rather incomparable as the efficiency of linear models at dealing with billions of such texts has proven less accurate and unscalable, while DNNs architectures perform on average 10\% better across classification tasks in NLP. 
CNN and LSTM networks are two popular DNN architectures. While CNN is an effective feature extractor, LSTM is suitable for modelling orderly sequence learning problems. CNN extracts word or character combinations and LSTM learns long-range word or character dependencies in texts. While each type of network has relative advantages, few works have explored combining both architectures into a single network~\cite{salminen2018anatomy}. {Conv-LSTM} is a robust architecture to capture long-term dependencies between features extracted by CNN and found more effective than structures solely based on CNN or LSTM~(where the class of a sequence depends on its preceding sequence). 

Further, various off-the-shelf word embeddings like Word2Vec, FastText, and GloVe, along with DNN classifiers i.e., CNNs, LSTMs, and GRUs, are employed in these approaches. Gomez et al.~\cite{gomez2020exploring} experimented with three different models including feature concatenation model~(FCM), spatial concatenation models~(SCM), and textual kernel model~(TKM). The basic FCM performed better than all, and they concluded that no additional improvement in performance could be obtained with the addition of images in hate speech detection. A multimodal approach by Blandfort et al.~\cite{blandfort2019multimodal} provides a promising improvement over the best single modality approach, yielding an average precision of 18\%. They use both early fusion and late fusion techniques in which a late fusion model with a stack of SVM classifiers showed a better performance. 

Transformer-based language models are becoming increasingly effective at NLP tasks and have made a notable shift in the performance on all text classification tasks including hate speech detection~\cite{sai2022explorative}. In addition, research has exposed the ability of BERT variants such as mBERT and XLM-RoBERTa to capture hateful context within social media content by using new fine-tuning methods based on transfer learning~\cite{sai2022explorative}. Sai et al.~\cite{sai2021towards} fine-tuned XLM-RoBERTa and mBERT for offensive speech detection in Dravidian languages and experimented with inter-language, inter-task, and multi-task transfer learning strategies leveraging resources for offensive speech detection in English. Ranasinghe et al.~\cite{ranasinghe2020multilingual} also fine-tuned XLM-RoBERTa for offensive language identification in three languages Bengali, Hindi, and Spanish, improving other deep learning models by fine-tuning Transformers~\cite{sai2022explorative}. 

Further, autoencoder~(AE)-based representation learning techniques have been employed to solve numerous supervised and unsupervised learning tasks. In AE architectures, weights of the encoder module are learned from both non-corrupted and unlabeled data. Subsequently, noisy supervised data with missing modalities is not suitable for learning latent features. Although, a simple AE can be used to reconstruct an output similar to the original input, it cannot handle multimodality. Therefore, AE-based multimodal representation learning techniques emerged, e.g., a new approach called deep orthogonal fusion~(DOF) model is proposed by Braman et al.~\cite{braman2021deep}. DOF first learns to combine information from multimodal inputs into a comprehensive multimodal risk score, by combining embeddings from each modality via attention-gated tensor fusion. To maximize the information gleaned from each modality, they introduce a new loss function called multimodal orthogonalization loss that increases model performance by incentivizing constituent embeddings to be more complementary. 

Patrick et al.~\cite{mmdcae} proposed another multimodal concept of learning called shared latent representation~(SLR) and latent representation concatenation~(LRC) techniques. Based on several studies covering text classification, sequence data, and imaging, they identified several limitations of SLR. First, the reconstruction loss for LRC is significantly lower compared to SLR-based representation learning technique. Secondly, when a classifier is trained on features learned by LRC, accuracy improves significantly, which is largely backed by lower reconstruction loss. 
On the other hand, accurate identification of hate speech in Bengali is a challenging task. Only a few restrictive approaches~\cite{romim2020hate,ishmam2019hateful,karim2020classification} have been proposed so far. Romim et al.~\cite{romim2020hate} prepared a dataset of 30K comments, making it one of the largest datasets for-a-kind of problem. However, this dataset has several issues. First, it is very imbalanced as the ratio of hate speech to non-hate speech is 10K:20K. Second, the majority of hate statements are very short in terms of length and word count compared to non-hate statements. Third, their approach exhibits a moderate level of effectiveness at identifying offensive or hateful statements, giving an accuracy of 82\%. 

Ismam et al.~\cite{ishmam2019hateful} collected hateful comments from Facebook and annotated 5,126 hateful statements. They classified them into six classes– hate speech, communal attack, insightful, religious hatred, political comments, and religious comments. Their approach, based on GRU-based DNN, achieved an accuracy of 70.10\%. In a recent approach~\cite{karim2020classification}, Karim et al. provided classification benchmarks for document classification, sentiment analysis, and hate speech detection for Bengali. Their approach is probably the first work among a few others on hate speech detection that combine fastText embedding with multichannel Conv-LSTM network. Since fastText works well with rare words such that even if a word was not seen during the training, it could be broken down into n-grams to get its corresponding embedding. Therefore, using fastText-based embeddings were more beneficial than Word2Vec and GloVe models. 

Therefore, instead of using manually engineered convolutional filters in CNN, convolutional and pooling layers can be stacked together to construct a stacked convolutional AE~(SCAE), to leverage better feature extraction capability. CAE learns more optimal filters by minimizing the reconstruction loss, which results in more abstract features from the encoder. This helps stabilize the pre-training, and the network converges faster by avoiding corruption in the feature space~\cite{guo2017deep}, making it more effective for very high dimensional data compared to vanilla AE-based multimodal learning. Therefore, considering the limitations of vanilla AE and SLR and the effectiveness of multimodal ML architectures, we constructed both MCAE based on a CAE and DOF. 

\section{Methods}\label{sec:methods}
In this section, we discuss our approach in detail, covering word embeddings and training of neural transformer architectures. 

\subsection{Data preprocessing} \label{subsec:preprocessing}
We remove HTML markups, links, image titles, special characters, and excessive use of spaces/tabs, before initiating the annotation process. Further, following preprocessing steps are followed before training ML and DNN baseline models:

\begin{itemize}
    \item \textbf{PoS tagging:} using BLSTM-CRF based approach~\cite{alam2016bidirectional}.
    \item \textbf{Removal of proper nouns:} proper nouns and noun suffixes were replaced with tags to provide ambiguity.
    \item \textbf{Hashtags normalization}: Hastags were normalized, with  the goal of supplying normalized hashtag content to  be  used  for the classification tasks. 
    \item \textbf{Stemming}: inflected words were reduced to their stem, base or root form. 
    \item \textbf{Stop word removal:} commonly used Bengali stop words are removed.
    \item Emojis, emoticons, and user mentions were removed.
    \item \textbf{Infrequent words}: tokens with low frequency~(TF and IDF) were removed. 
\end{itemize}

As BERT-based models perform better classification accuracy on uncleaned texts, we did not perform major preprocessing tasks, except for the lightweight preprocessing discussed above. On the other hand, for the imaging modality, we performed some preprocessing as the dataset has varying sizes. Therefore, images are converted into RGB format before extracting features, while the sizes are adjusted according to deep CNN model specification. 

\subsection{Neural word embeddings}
We train the \emph{fastText}~\cite{fastText} word embedding model on Bengali articles used for the classification benchmark study by Karim et al.~\cite{karim2020classification}. The preprocess reduces vocabulary size due to the colloquial nature of the texts and some degree, addresses the sparsity in word-based feature representations. We have also tested, by keeping word inflexions, lemmatization, and lower document frequencies\footnote{\scriptsize{We observe slightly better accuracy using lemmatization, hence we report the result based on it.}}.

The fastText model represents each word as an n-gram of characters, which helps capture the meaning of shorter words and allows the embeddings to understand suffixes and prefixes. Each token is embedded into a 300-dimensional vector space, where each element is the weight for the dimension for that token. Since the annotated hate statements are relatively short, we constrain each sequence to 100 words by truncating longer texts and pad shorter ones with zero values to avoid padding in convolutional layers with many blank vectors. 

\subsection{Training of DNN baseline models}
We train three DNN baselines: CNN, Bi-LSTM, and Conv-LSTM. Weights of the embedding layer for each network are initialized with the embeddings based on the fastText embedding model. The embedding layer maps each hate statement into a \emph{sequence}~(for LSTM and CNN layers) and transforms it into feature representation, which is then flattened and fed into a fully connected softmax layer. Further, we add Gaussian noise and dropout layers to improve model generalization. AdaGrad optimizer is used to learn the model parameters by reducing the categorical-cross-entropy loss. We train each DNN architecture 5 separate times in a 5-fold CV setting, followed by measuring the average F1-score on the validation set to choose the best hyperparameters. 

\subsection{Training of transformer-based models}
\begin{figure*}
    \centering
    \includegraphics[width=0.6\textwidth]{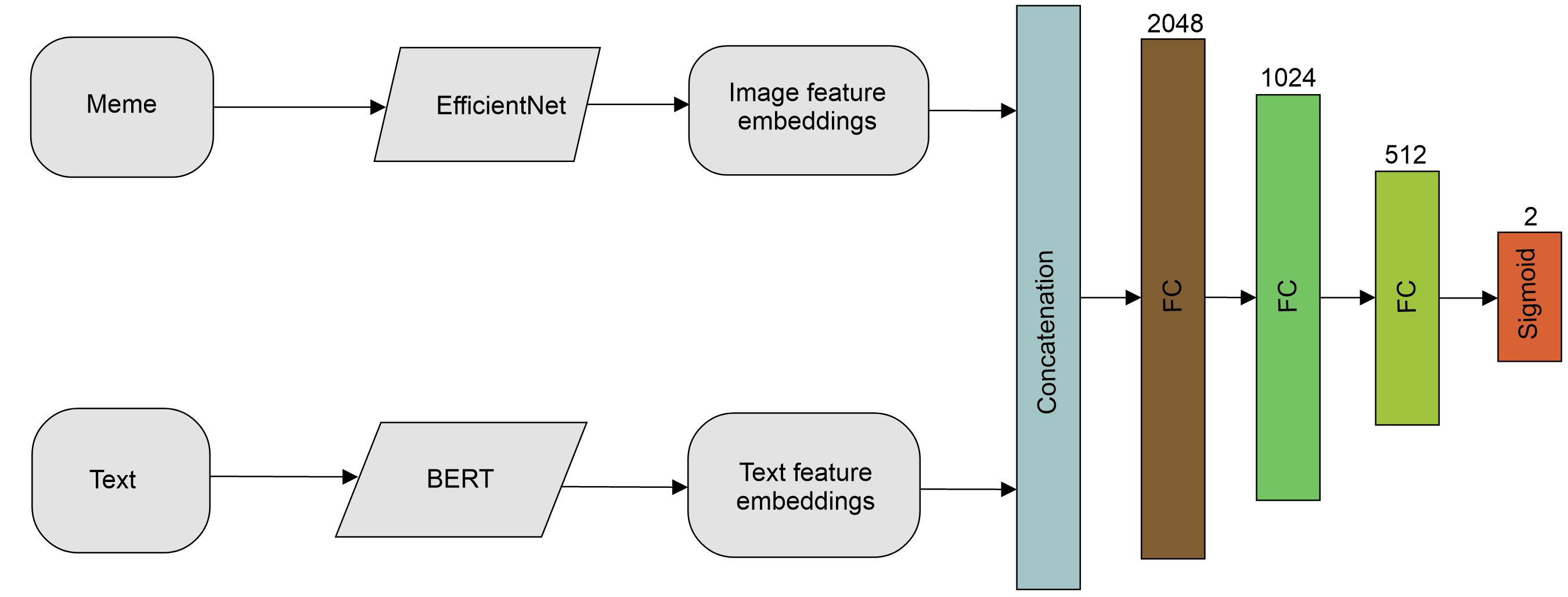}   
    \caption{Schematic representation of the approach for hate speech detection}    
    \label{fig:network_architecture}
\end{figure*}

As shown in \cref{fig:network_architecture}, we train monolingual Bangla BERT-base, mBERT, and XLM-RoBERTa large models. Bangla-BERT-base\footnote{\scriptsize{\url{https://huggingface.co/sagorsarker/bangla-bert-base}}} is a pretrained Bengali language model built with BERT-based mask language modelling. RoBERTa~\cite{liu2019roberta} is an improved variant of BERT, which is optimized by setting larger batch sizes, introducing dynamic masking, and training on larger datasets. XLM-RoBERTa~\cite{conneau2019unsupervised} is a multilingual model trained on web-crawled data. XLM-RoBERTa not only outperformed other transformer models on cross-lingual benchmarks but also performed better on various NLP tasks in low-resourced language settings~\cite{sai2021towards}.

\begin{figure*}[h]
	\centering
	\includegraphics[width=0.5\textwidth]{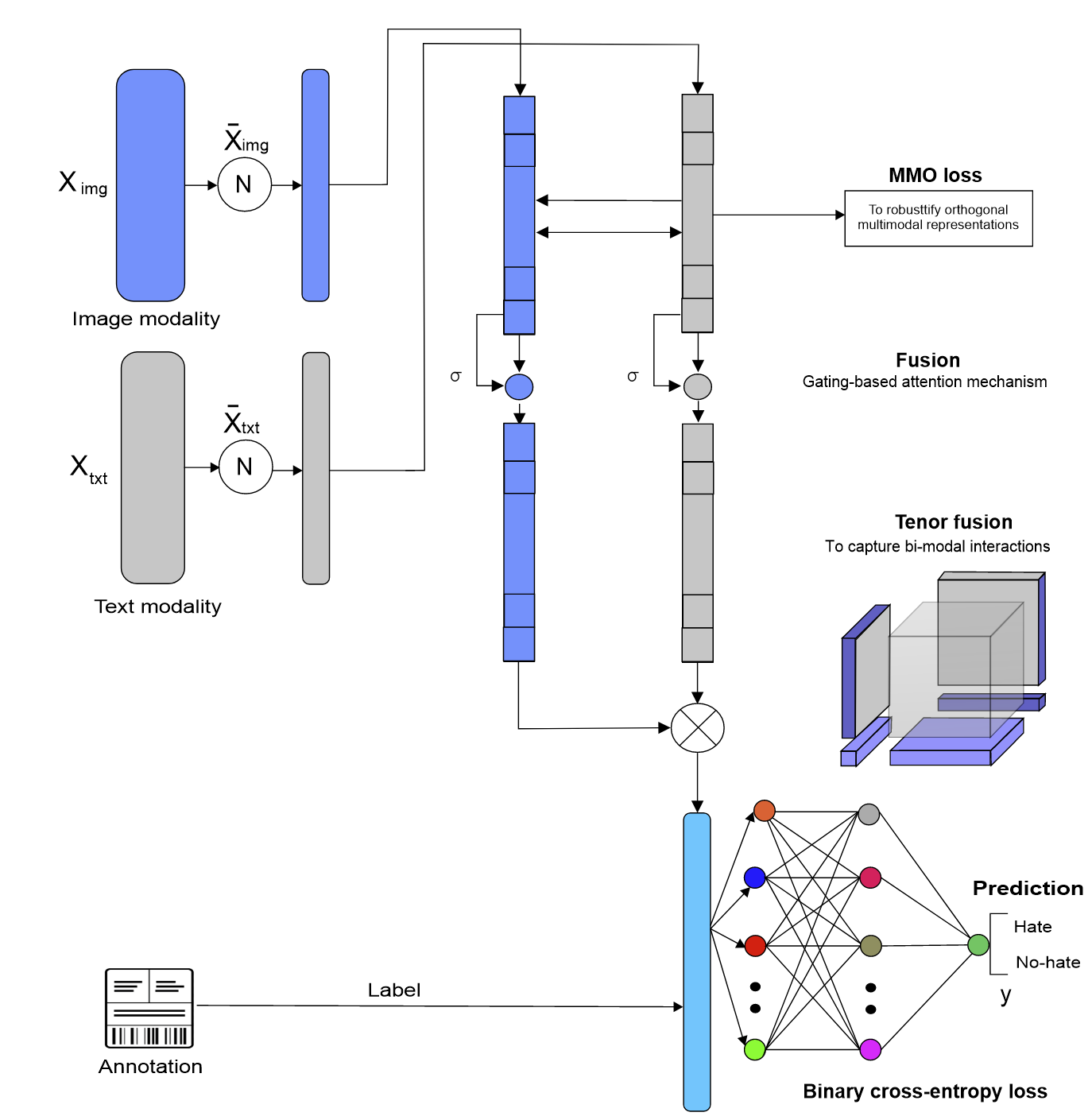}
	\caption{DOF model for hate speech detection~(recreated based on literature~\cite{braman2021deep})} 
	\label{fig:dof}
\end{figure*}

We shuffle training data for each epoch and apply gradient clipping to clip error derivative to a threshold during backward propagation of the network. We set the initial learning rate to $2 e^{-5}$ and employ Adam optimizer with the scheduled learning rate. PLMs are fine-tuned by setting the maximum input length to $256$. We experimented with 2, 3, and 4 layers of multi-head attention, followed by placing a fully connected softmax layer on top. Several experiments are carried out with different hyperparameters combinations, as shown in \cref{tab:hyperparams}. 

\begin{table}
    \centering
    \caption{Hyperparams for language models}
    \vspace{-4mm}
    \begin{tabular}{|c|c|} 
    \hline
      \textbf{Hyper-parameters} & \textbf{Value}\\
      \hline
      Learning-rate & \{2e-5, 3e-5\} \\ \hline 
      Epochs &  \{5, 6, 10, 20\} \\ \hline 
      Max sequence length & \{64, 128, 256\} \\ \hline 
      Dropout & \{0.1, 0.2, 0.3, 0.5\} \\ \hline 
      Batch size & \{16, 32, 64, 128\} \\
      \hline
    \end{tabular}
    \label{tab:hyperparams}
\end{table}

\subsection{Multimodal fusion and classification}
Let $X$ be a training minibatch of data for $N$ samples, each containing $M$ modalities such that $X=\left[x_{1}, x_{2}, \ldots, x_{M}\right]$, where $x_{m}$ represents the data for respective modality $m$. We employ two approaches to learn the joint embeddings~(LRC and late fusion using DFO), followed by supervised-fine tuning for the classification. 

In LRC, first image, image-text, and textual embedding are first learned. Assuming input $x_{m} \in \mathbb{R}^{D}$ for each of $m \in \mathbb{R}^K$ modalities is consisting of $N$ samples. A convolutional layer of CAE calculates the convolutional feature map. Since individual latent representation is required to have the same dimensionality~\cite{mmdcae}, we generate a combined representation for all input modalities, instead of one latent representation for each input modality. Max-pooling operation is then performed, which downsamples the output of the convolutional layer by taking the maximum value in each non-overlapping sub-region. Thus, $x_{m}$ is mapped and transformed into a lower-dimensional embedding space $z_m$. The latent-space representation $z_m=g_{\phi}(x_{m})$ is learned in the bottleneck layer~\cite{mmdcae}: 

\begin{equation}
    z_m = h_{m}=g_\phi \left({x}_{m}\right)=\sigma\left(W_{m} \oslash x_{m}+b_{m}\right),
    \label{eq:fcuk_1}
\end{equation}

where the encoder is a sigmoid function $g(.)$ parameterized by $\phi$, while the decoder function $f(.)$ is parameterized by $\Theta$. The final feature maps $Z_m$ are latent variables, specific to modality $m$. In \cref{eq:fcuk_1}, where $\phi$ are trainable parameters~(including a weight matrix $W_{k} \in \mathbb{R}^{p \times q}$ and a bias vector $b_{m} \in \mathbb{R}^{q}$ specific to respective modality $m$, where $p$ and $q$ are the numbers of input and hidden units), $\oslash$ is the convolutional operation, and $\sigma$ is the exponential linear unit activation function. The decoder reconstructs the original input $X_{m}$ from the latent representation $Z_m$ using function $f(.)$. The hidden representation $h_{m}$ is mapped back to reconstructed version ${x}_{m}$, similar to original input ${X}_{k}$~\cite{mmdcae}: 

\begin{equation}
    \hat{x_m}=f_{\theta}\left(z_m\right)=f_{\theta}\left(g_{\phi}({x_m})\right) \label{eq:mcae_eq_1},
\end{equation}

where parameters $(\theta,\phi)$ are jointly learned to reconstruct the original input. As this is analogous to learn an identity function, such that ${x_m} \approx f_{\theta}\left(g_{\phi}({x_m})\right)$,  $f_{\theta}\left(g_{\phi}({x_m})\right)$ is equivalent to $\Psi \left(\hat W_m * h_m + \hat b_{m}\right)$, which yields \cref{eq:mcae_eq_1} into: 

\begin{equation}
    \hat{x_m}=\Psi \left(\hat W_m \odot h_m + \hat b_{m}\right),
\end{equation}

where $\odot$ is the transposed convolution operation, $\theta$ are trainable parameters~i.e., weight matrix $\hat W_{m} \in \mathbb{R}^{n \times p}$, bias vector $\hat b_{m}$) specific to modality $m$, and sigmoid activation function $\Psi$. Let $x_t$ and $x_i$ be the text and image modalities, then each $x_m$ is transformed into following hidden representations~\cite{mmdcae}.

\begin{equation}
    \begin{array}{l}
        {h_{t}=\sigma \left(W_{t} \oslash x_{t}+b_{t}\right)} \\
        {h_{i}=\sigma \left(W_{i} \oslash x_{i}+b_{i}\right)},
    \end{array}
\end{equation}  

where $\left\{W_{t}, W_{i}\right\}$ are encoder's weight matrices, $\left\{b_{t} and b_{i}\right\}$ are bias vectors for the text and image modalities, respectively. Last element of the hidden dimension is the dimensionality of the modality-specific latent representation. The mean squared error is used as the reconstruction loss:  

\begin{equation}
    L_{\mathrm{m}}(\theta, \phi)=\frac{1}{n} \sum_{i=1}^{n}\left({x_m}-f_{\theta}\left(g_{\phi}\left({x_m}\right)\right)\right)^{2} +\lambda\left\|W_{m}\right\|_{2}^{2} \cdot
\end{equation} 

By replacing $f_{\theta}\left(g_{\phi}\left({x_m}\right)\right)$ with $\hat{x}_{m}$, above equation yields: 

\begin{equation}
    L_{\mathrm{m}}(\theta, \phi)=\frac{1}{n} \sum_{i=1}^{n}\left({x_m}-\hat{x}_{m}\right)^{2} +\lambda\left\|W_{m}\right\|_{2}^{2},
\end{equation} 

where $\lambda$ is the activity regularizer and $W_{m}$ is network weights specific to input modality $m$. In the cross-modality stage, a concatenation layer concatenates individual latent representations $h_{t}$ and $h_{i}$ into following single representation:  

\begin{equation}\label{eq:concat_vector}
    h_{mcae}=\sigma\left(W_{mcae}\left[h_{t} \oplus h_{i} \right]+b_{mcae}\right),
\end{equation}

where $\oplus$ signifies concatenation. As of $DOF$ model, we customize it in a classification setting. First, a trainable unimodal network is trained that takes $x_{m}$ as input and generates deep embeddings~\cite{braman2021deep}:

\begin{equation}
    h_{m}=\Phi_{m}\left(x_{m}\right) \in \mathbb{R}^{l_{1} x N},
\end{equation}

where $\Phi_{m}$ trainable unimodal network. When $M>1$, embeddings from each modality are combined in a multimodal fusion network. For each $h_{m}$, an attention mechanism is applied to control its expressiveness based on information from other modalities. An additional fully connected layer results in $h_{m}^{S}$ of length $l_{2}$. Attention weights of length $l_{2}$ are obtained through a bi-linear transformation of $h_{m}$ with all other embeddings (denoted as $H_{pr}$ ), then applied to $h_{m}^{S}$ to yield the following attention-gated embedding~\cite{braman2021deep}: 

\begin{equation}
    h_{m}^{*}=a_{m} * h_{m}^{S}=\sigma\left(h_{m}^{T} * W_{A} * H_{pd}\right) * h_{m^{*}}^{S}
\end{equation}

To capture all possible interactions between modalities, we combine attention-weighted embeddings through an outer product between modalities, known as tensor fusion. A value of 1 is also included in each vector, allowing for partial interactions between modalities and for the constituent unimodal embeddings to be retained. The output matrix is the following $M$-dimensional hypercube of all multimodal interactions with sides of length $l_{2}+1$~\cite{braman2021deep}: 

\begin{equation}
    F=\left[\begin{array}{c}
    1 \\
    h_{1}^{*}
    \end{array}\right] \otimes\left[\begin{array}{c}
    1 \\
    h_{2}^{*}
    \end{array}\right] \otimes \ldots \otimes\left[\begin{array}{c}
    1 \\
    h_{M}^{*}
    \end{array}\right]
\end{equation}

\Cref{fig:dof} depicts the schematic representation of multimodal fusion of memes and texts. It contains sub-regions corresponding to unaltered unimodal embeddings, pairwise fusions between 2 modalities. A final set of fully connected layers, denoted by $\Phi_{F}$, is applied to tensor fusion features for a final fused embedding $h_{F}=\Phi_{F}(F)$~\cite{braman2021deep}. Unimodal embeddings before the fusion level should be orthogonal s.t. each modality contributes unique information to outcome prediction, rather than relying on signal redundancy between modalities~\cite{braman2021deep}. Inspired by this, we updated each $\Phi_{m}$ through MMO loss to yield embeddings that better complement other modalities. For the set of embeddings from all modalities $H \in \mathbb{R}^{l_{1} x M * N}$, the MMO loss is computed as follows~\cite{braman2021deep}: 

\begin{equation}
    L_{MMO}=\frac{1}{M * N} \sum_{m=1}^{M} \max \left(1,\left\|h_{m}\right\|_{*}\right)-\|H\|_{+}
\end{equation}

where $\|\cdot\|$ denotes the matrix nuclear norm. As the loss is the difference between the sum of nuclear norms per embedding and the nuclear norm of all embeddings combined, it penalizes the scenario where the variance of two modalities separately is decreased when combined and minimized when all unimodal embeddings are fully orthogonal.Having the multimodal embeddings, we perform supervised fine-tuning by optimizing the binary cross-entropy loss. 

\section{Experiment Results}\label{sec:exp}
We discuss experimental results both qualitatively and qualitatively. Besides, we provide a comparative analysis with baselines.

\subsection{Datasets}\label{section:3}
We extend the \emph{Bengali Hate Speech Dataset}~\cite{karim2020classification} with 4,500 labelled memes, making it the largest and only multimodal hate speech dataset in Bengali to date. We follow a bootstrap approach for data collection, where specific types of texts containing common slurs and terms, either directed towards a specific person or entity or generalized towards a group, are only considered. Texts and memes were collected from Facebook, YouTube comments, and newspapers. While the ``Bengali Hate Speech Dataset'' categorized observations into political, personal, geopolitical, religious, and gender abusive hates, we categorized them into hateful and non-hateful, keeping their respective contexts intact.  

Further, to reduce possible bias, unbiased contents are supplied to the annotators and each label was assigned based on a majority voting on the annotator's independent opinions\footnote{\scriptsize{A linguist, a native speaker \& an NLP researcher participated in annotation process.}}. To evaluate the quality of the annotations and to ensure the decision based on the criteria of the objective, we measure inter-annotator agreement w.r.t \emph{Cohen's Kappa} statistic~\cite{chen2005macro}. 

\subsection{Experiment setup}
Methods\footnote{\scriptsize{{\url{https://github.com/rezacsedu/Multimodal-Hate-Bengali}.}}} were implemented in \emph{Keras} and \emph{PyTorch}. Networks are trained on Nvidia GTX 1050 GPU. Each model is trained on 80\% of data~(of which 10\% is used as validation), followed by evaluating the model on 20\% held-out data. We report precision, recall, F1-score, and \emph{Matthias correlation coefficient}~({MCC}). 

\begin{figure*}
	\centering
	\begin{subfigure}{\linewidth}
		\centering
		\includegraphics[width=0.7\textwidth]{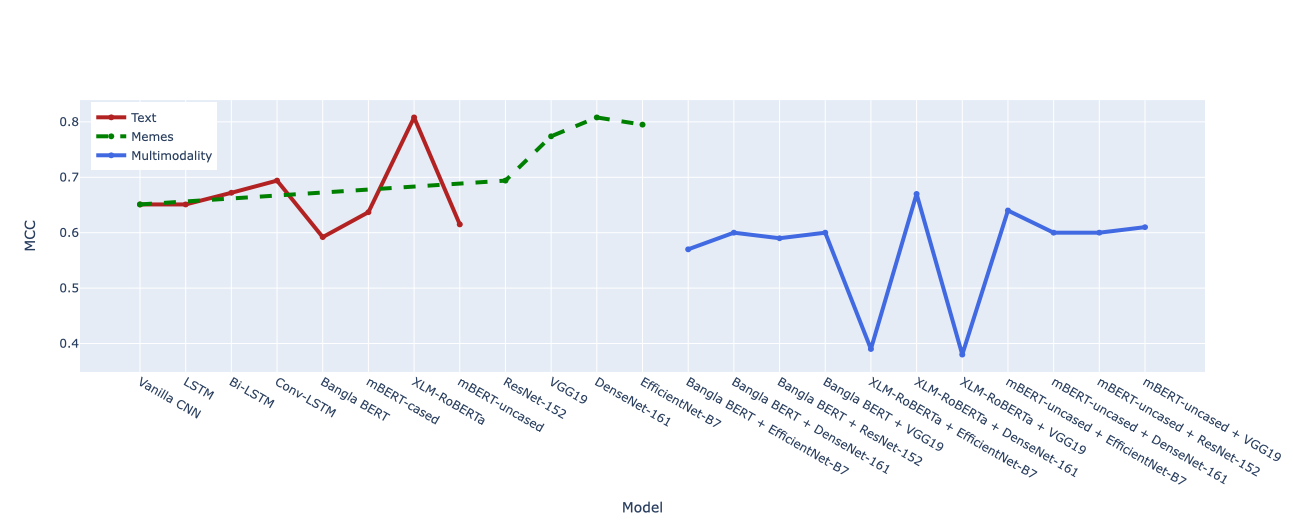}
		\caption{MCC scores across modalities}
        \label{fig:probing}
	\end{subfigure}\vfill
	\begin{subfigure}{\linewidth}
		\centering
		\includegraphics[width=0.7\textwidth]{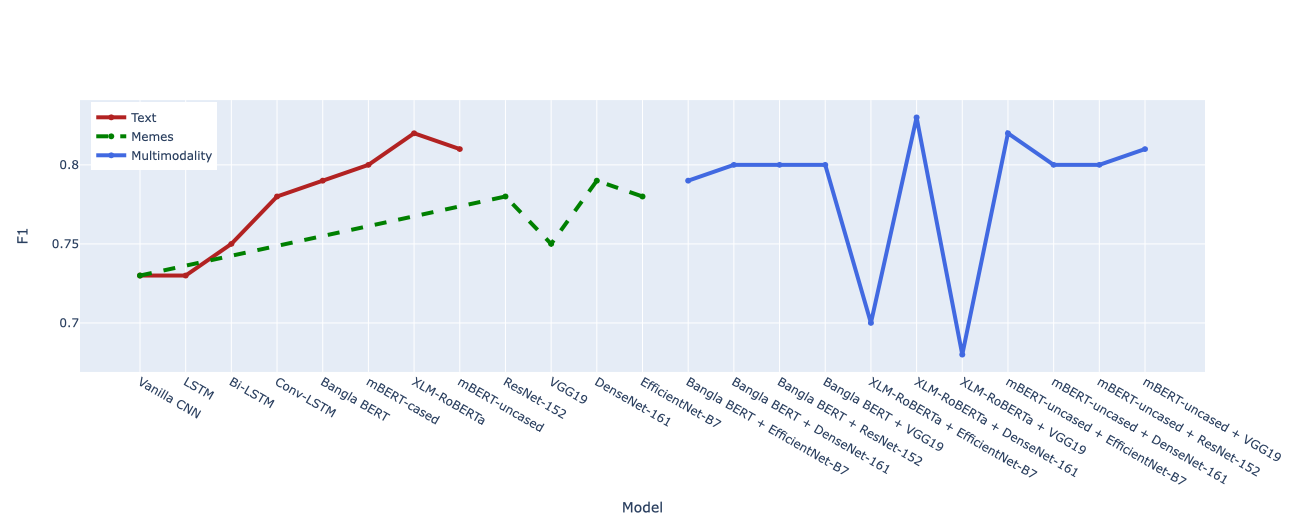}
		\caption{F1 scores across modalities}
        \label{fig:purturbing}
	\end{subfigure}
	\caption{Effects of individual modalities w.r.t MCC and F1 scores} 
	\label{fig:mcc_vs_f1_scores}
\end{figure*}

\subsection{Analysis of hate speech detection}
We report modality-specific results. Classification results based on text modality are shown in \cref{table:hate_result_v2}. We evaluated DNN or ML baseline and 4 variants of BERT on held-out test set and report the result. We observed that single text modality has comparable performance to DNN baseline models. Conv-LSTM outperformed other DNN baselines, giving an F1 score of 78\%, which is 4\% to 6\% better other DNN models. In terms of MCC~(i.e., 0.69), Conv-LSTM model is about 2\% to 4\% better than other DNN baselines. 

\begin{table*}[h]
            \centering
            \caption{Hate speech detection based on texts}
            \label{table:hate_result_v2}
            \vspace{-4mm}
            \scriptsize{
            \begin{tabular}{c|l|l|l|l|l}
            \hline
            \textbf{Method} & \textbf{Classifier} & \textbf{Precision} & \textbf{Recall} & \textbf{F1} & \textbf{MCC}\\ \hline
             \multirow{3}{*}{DNN baselines}
             & Vanilla CNN & 0.74 &    0.73 &    0.73 &    0.651\\
             & LSTM & 0.74 &    0.73 &    0.73 &    0.651\\
             & Bi-LSTM & 0.75 &    0.75 &    0.75 &    0.672\\
             & Conv-LSTM & \textbf{0.79} &\textbf{0.78} &    \textbf{0.78} & \textbf{0.694}\\
             \hline
            \multirow{5}{*}{BERT variants} & Bangla BERT & 0.80  &    0.79  &  0.79 &    0.592\\
             & mBERT-cased & 0.80  &    0.80   &   0.80 &    0.637 \\
             & XLM-RoBERTa & \textbf{0.82}  &   \textbf{0.82 } & \textbf{0.82}  &   \textbf{0.808} \\
             & mBERT-uncased & 0.81    &  0.81   &   0.81  &    0.615 \\ 
             \hline
            \end{tabular}
            }
\end{table*}

\begin{table*}[h]
            \centering
            \caption{Hate speech detection based on memes}
            \label{table:hate_result_v1}
            \vspace{-4mm}
            \scriptsize{
            \begin{tabular}{c|l|l|l|l|l}
            \hline
            \textbf{Method} & \textbf{Classifier} & \textbf{Precision} & \textbf{Recall} & \textbf{F1} & \textbf{MCC}\\ \hline
             \multirow{3}{*}{CNN baselines}
             & Vanilla CNN & 0.74 &    0.73 &    0.73 &    0.651\\
             & ResNet-152 & \textbf{0.79} &\textbf{0.78} &    \textbf{0.78} & \textbf{0.694}\\
             & VGG19 & 0.75  &    0.75   &   0.75 &    0.774 \\
             & DenseNet-161 & \textbf{0.79}  &   \textbf{0.79 } & \textbf{0.79}  &   \textbf{0.808} \\
             & EfficientNet-B7 & 0.79    &  0.78   &   0.78  &    0.795\\
             \hline
            \end{tabular}}
\end{table*}

\begin{table*}[h]
            \centering
            \caption{Hate speech detection for multimodality}
            \label{table:hate_result_v3}
            \vspace{-4mm}
            \scriptsize{
            \begin{tabular}{c|l|l|l|l|l}
            \hline
            \textbf{Method} & \textbf{Classifier} & \textbf{Precision} & \textbf{Recall} & \textbf{F1} & \textbf{MCC}\\ \hline
            \multirow{12}{*}{ConvNet + BERTs} &
             Bangla BERT + EfﬁcientNet-B7 & 0.79 & 0.79 & 0.79 & 0.57 \\ &
             Bangla BERT + DenseNet-161 & \textbf{0.80} & \textbf{0.80} & \textbf{0.80} & \textbf{0.60} \\ &
             Bangla BERT + ResNet-152 & 0.80 & 0.80 & 0.80 & 0.59 \\ &
             Bangla BERT + VGG19 & \textbf{0.80} & \textbf{0.80} & \textbf{0.80}  & \textbf{0.60} \\ &
             
             XLM-RoBERTa + EfﬁcientNet-B7 & 0.70 & 0.70 & 0.70 & 0.39 \\ &
             XLM-RoBERTa + DenseNet-161 & \textbf{0.84} & \textbf{0.83} & \textbf{0.83} & \textbf{0.67} \\ &
             XLM-RoBERTa + ResNet-152 & 0.73 & 0.72 & 0.72 & 0.44 \\ &
             XLM-RoBERTa + VGG19 & 0.70 & 0.68 & 0.68  & 0.38 \\ &
             
             mBERT-uncased + EfﬁcientNet-B7 & \textbf{0.82} & \textbf{0.82} & \textbf{0.82} & \textbf{0.64} \\ &
             mBERT-uncased + DenseNet-161 & 0.80 & 0.80 & 0.80 & 0.60 \\ &
             mBERT-uncased + ResNet-152 & 0.80 & 0.80 & 0.80 & 0.60 \\ &
             mBERT-uncased + VGG19 & 0.81 & 0.81 & 0.81  & 0.61 \\
             \hline
            \end{tabular}}
\end{table*}

As shown in \cref{table:hate_result_v2}, XLM-RoBERTa outperformed all transformer models and turns out to be the best model, yielding an F1 score of 82\%, which is 2\% $\approx$ 3\% better than other transformer models. mBERT model~(cased and uncased) performed moderately well than Bangla BERT. Results based on unimodal visual models are reported in \cref{table:hate_result_v1}. Among trained CNN architectures~(i.e., ResNet-152, VGG19, DenseNet-161, EfficientNet-B7), DenseNet-161 achieved an F1-score of 79\% and MCC score of 0.808, which is about 4\% $\approx$ 6\%~(w.r.t F1) or 2\% $\approx$ 15\%~(w.r.t MCC) better than that of other architectures.

We experimented with different combinations of multimodal models for textual feature extraction and vision modality. We report the results of top-4 models in \cref{table:hate_result_v3}. As shown, the multimodal fusion of mBERT-uncased + EfficientNet-B7 yielding F1-scores of 0.82 outperforms all mBERT + ConvNets fusion. At the same time, Bangla BERT + DenseNet-161 and Bangla BERT + VGG19 fusion combinations, which are the best performing models in Bangla BERT + ConvNets architectures by 2\% w.r.t F1-score. On the other hand, the multimodal fusion of XLM-RoBERTa + DenseNet-161 turned out to be the best performing model w.r.t precision and F1-score, yielding the highest MCC score of 0.67, which is about 3 to 15\% better than other multimodal fusions. 

\section{Conclusion}\label{sec:con}
In this paper, we proposed hate speech detection for under-resourced Bengali language in multimodal setting that jointly analyze textual and visual information for hate speech detection. Our study suggests that: i) feature selection can have non-trivial impacts on learning capabilities of ML and DNN models, ii) texts are mot useful modality for hate speech identification, while memes are moderately useful. However, none of multimodal models outperform unimodal models analyzing only textual data, as shown in \ref{fig:mcc_vs_f1_scores}. We believe that our computational resources~(annotated dataset, language models, source codes, and interpretability techniques) will further advance NLP research for the under-resourced Bengali language. 

Our approach has several limitations too. First, we had a limited amount of labelled data at hand during the training. Secondly, our approach is like black-box model, thereby not having the capability to explain and reason the decision why a certain meme and associate text contain hate. In the future, we want to overcome above limitations by: i) extending the datasets with a substantial amount of samples, ii) improving explainability of the model by employing different interpretable ML techniques in order to provide global and local explanations of the predictions in a post-hoc fashion and measures of explanations w.r.t faithfulness.  Besides, we want to focus on other interesting areas such as named entity recognition, part-of-speech tagging, and question answering. 

\printbibliography

\end{document}